\documentclass[a4paper,fleqn]{cas-dc}

\usepackage[numbers,sort&compress]{natbib}

\def\tsc#1{\csdef{#1}{\textsc{\lowercase{#1}}\xspace}}
\tsc{WGM}
\tsc{QE}
\tsc{EP}
\tsc{PMS}
\tsc{BEC}
\tsc{DE}

\begin{document}
\let\WriteBookmarks\relax
\def\floatpagepagefraction{1}
\def\textpagefraction{.001}
\pagestyle{empty}

\title [mode = title]{A Survey on Open Set Recognition}                      

\author{Atefeh Mahdavi}
\ead{amahdavi@fit.edu}
\cormark[1]
\author{Marco Carvalho}
\ead{mcarvalho@cs.fit.edu}
\cortext[cor1]{Corresponding author}
\address{College of Engineering and Science, Florida Institute of Technology, Melbourne, FL 32901, United States}

\begin{abstract}
Open Set Recognition (OSR) is about dealing with unknown situations that were not learned by the models during training. In this paper, we provide a survey of existing works about OSR and distinguish their respective advantages and disadvantages to help out new researchers interested in the subject. The categorization of OSR models is provided along with an extensive summary of recent progress. Additionally, the relationships between OSR and its related tasks including multi-class classification and novelty detection are analyzed. It is concluded that OSR can appropriately deal with unknown instances in the real-world where capturing all possible classes in the training data is not practical. Lastly, applications of OSR are highlighted and some new directions for future research topics are suggested.
\end{abstract}

\begin{keywords}
classification \sep open set recognition \sep multi-task learning \sep support vector machines \sep risk of the unknown \sep
\end{keywords}
\maketitle

\section{Introduction}
In OSR, only a limited number of known classes are available at the time of training the model and the possibility of unknown classes never seen at training time emerges in the test environment. In such a setting, the unknown classes and their risk should be considered in the algorithm. Such systems require not only to identify and discriminate instances that belong to the source domain (i.e., the seen known classes contained in the training dataset) but also to reject unknown classes in the target domain (classes used in the testing phase). Until recently, the success of almost all machine-learning-based systems has been obtained by conducting them on \textquotedblleft closed-set\textquotedblright classification tasks. In such systems, the source and target domains are assumed to contain the same object classes and the system is only tested on known classes that have been seen during training. Different from the \textquotedblleft closed set\textquotedblright setting, a more realistic scenario is solving real-world problems consisting of an \textquotedblleft open set\textquotedblright of objects. With the advent of building intelligent systems and utilizing machine-learning-based systems, a wide range of applications require robust AI methods. Handling the \textquotedblleft unknown unknowns\textquotedblright \cite{dietterich2017steps} can be considered as one of the approaches that enable the system to act robustly in the face of limitations and unmodeled aspects of the world. Ignoring unknown objects causes improper development of the systems and limits their usability. However, building a correct and complete model for the recognition/classification task in the real dynamic world poses multiple challenges as anticipating and training all possible examples of unknown objects are prohibitive and the model may fail when assessed in testbeds. 
In the past literature, authors have proposed terms such as \textquotedblleft open set recognition\textquotedblright  \cite{bendale2016towards}, \textquotedblleft open category learning\textquotedblright  and \textquotedblleft open-world recognition\textquotedblright  \cite{bendale2015towards} that can respond to model failure. A resurgence of interest in solving this challenging task has also led researchers to propose a set of related topics under different learning paradigms. These studies include domain adaptation \cite{patel2015visual, csurka2017domain}, transfer learning \cite{pan2009survey, weiss2016survey, tan2018survey} and few/zero-shot learning \cite{wang2019survey, snell2017prototypical, fu2018recent, fei2006one}. In this paper, we will use the term open set recognition.
\par 
There are two broad categories of OSR systems. The first one refers to the task of discriminating known class instances from unknown class instances. This mechanism which is not able to distinguish between the known classes acts as a detector rather than a classifier. This technique is applied in research such as \cite{bodesheim2013kernel, scheirer2012toward}. In the second category, in which the number of classes is more than two, OSR is concerned with distinguishing between the known classes. This system identifies unknowns and labels the input as one of the known classes it best fits or as unknown \cite{ ge2017generative, jain2014multi, bendale2015towards, bendale2016towards}. A challenge faced by a potential solution to the OSR is estimating the correct probability of all known classes and maintaining the performance on them, along with a simultaneous precise prediction of unknown classes and optimizing the model for them. 
\par
This paper is organized as follows: First, we briefly differentiate OSR from multi-class classification and novelty detection problems and discuss their limitations. Then, we propose two broad categories for OSR algorithms and details of the important studies under each category. The categories described here encompass statistical-based and deep-neural-network-based algorithms. {Table~\textcolor{blue}{\ref{tab0}}} represents the classification of these categories and their sub-categories formed by the existing methods. In the last section, we provide an overall conclusion for this review.

\begin{table*}[width=1.9\textwidth]
\caption{A global picture of the existing OSR methods. }
\resizebox{\textwidth}{!}{%
\begin{tabular}{@{}rrrrcrrr@{}}\toprule
& \multicolumn{2}{c}{\textbf{Statistical Models}} &
\phantom{abc} & \multicolumn{4}{c}{\textbf{Deep Neural Networks}}\\
\cmidrule{1-4} \cmidrule{6-8}
\textbf{Rejection-adapted SVM} & \textbf{Sparse Representation} & \textbf{ Distance-based} & \textbf{Margin Distribution} && \textbf{Adversarial Learning} & \textbf{Background-class-based Modeling} & \textbf{Others}\\  \midrule

\cite{scheirer2012toward} \cite{jain2014multi} \cite{scheirer2014probability} & \cite{zhang2016sparse} & \cite{bendale2015towards} \cite{de2016online} \cite{junior2017nearest} \cite{doan2017overcoming} 
 & \cite{rudd2017extreme} \cite{vignotto2018extreme} && \cite{ge2017generative} \cite{neal2018open} \cite{jo2018open} & \cite{dhamija2018reducing} \cite{ren2015faster} \cite{zhang2017towards} \cite{liu2016ssd}& \cite{bendale2016towards} \cite{yoshihashi2019classification} \cite{cardoso2015bounded} 
\\
\cite{junior2016specialized} \cite{scherreik2016open} \cite{panareda2017open} &  & &  && \cite{yu2017open}\cite{yang2019open} \cite{saito2018open}& & \cite{cardoso2017weightless} \cite{shu2017doc}\cite{oza2019c2ae}
\\
\cite{da2014learning} \cite{liu2019separate} \cite{pritsos2018open}&  &  &  && & & \cite{oza2019deep}\cite{lian2019known}\cite{hassen2018learning}\\

\bottomrule
\label{tab0}
\end{tabular}}
\end{table*}

\section{OSR vs. Multi-class Classification and Anomaly Detection}
\begin{figure*}[ht]
    \centering
       \includegraphics[width=17cm]{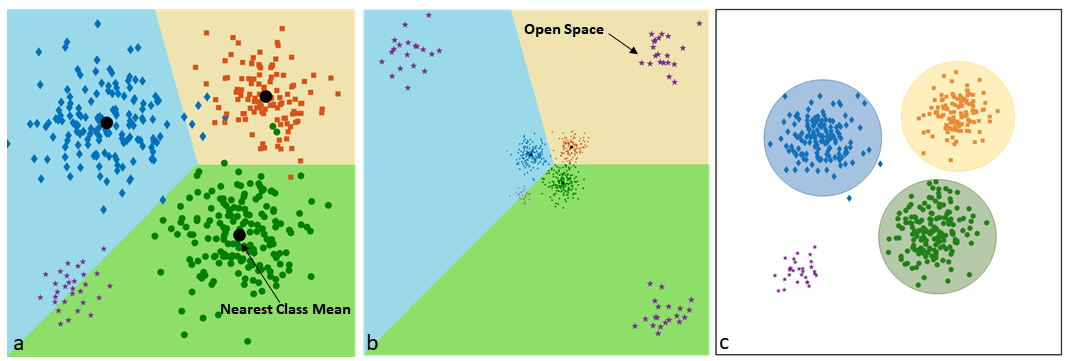}
    \vspace{-0.7em}
    \caption{An overview of the issue of OSR}
    \label{fig2}
\end{figure*}

OSR is referred to as a classification-based task. Most of the approaches to OSR were formed based on regular classifiers due to their closeness to the classification task; however, the adaptation of a classifier which is valid for OSR is not always possible. Classification and anomaly detection are the closest relatives to OSR. The relationship between OSR and these related areas is summarized in {Table~\textcolor{blue}{\ref{tab1}}}.
In a conventional multi-class classifier, because of the closed set assumption, all inputs are labeled and classified into one of the known classes observed during training. More precisely, in the closed set classification task, the learner only has access to a fixed set of known classes $C= \left \{ L_{1}, L_{2},...L_{M} \right \}$ and constructs an M-class classifier during the training phase. The resulting classifier is tested on the data from only the M classes. However, a problem emerges with the appearance of a test sample from an unknown class which does not belong to any of the known classes. Thus, the most likely class for an input observation is always provided and an unknown will wrongly be recognized as a sample belonging to one of those pre-defined classes. In OSR; however, knowledge of the entire set of possible classes cannot be considered during training. The classifier is allowed to predict classes from the set of $C^{'}= \left \{ L_{1}, L_{2},...,L_{M},L_{M+1},...,L_{M+\Omega} \right \}$, where classes $L_{M+1}$ through $L_{M+\Omega}$ cover all unknown classes not observed during training but which appeared at query time. A test sample may be predicted to belong either to one of the known classes $c_{i} \in C$ or to an unknown one. The difference between OSR and traditional classification is visualized in {Figure~\textcolor{blue}{\ref{fig2}}}. The decision boundaries in {Figure~\textcolor{blue}{\ref{fig2}}(a)} are considered by training a traditional Nearest Class Mean (NCM) classifier on three different known classes illustrated by diamonds, circles, and squares and the unknown inputs represented by stars. {Figure~\textcolor{blue}{\ref{fig2}}(b)} demonstrates the distribution of original dataset in the open space when zooming out from the closed three-class model. Having incomplete knowledge of the entire set of possible classes, this classifier assigns class labels from the closed training set to an unlimited region. Therefore, at the classification time, the unknown inputs in the open space will be misclassified. On the other hand, OSR discriminates known samples and limits the scope of decisions by the support of the training data (see {Figure~\textcolor{blue}{\ref{fig2}}(c)}). 
\par

\begin{table*}[width=1.9\textwidth]
  \caption{Relationship between OSR and the related areas}
  \label{tab:commands}
  \resizebox{\textwidth}{!}{%
  \begin{tabular}{cccl}
    \toprule
    Settings & Training Data & Testing Data & Tasks\\
    \midrule
    \texttt{Traditional Classification}& Known & Known & Classifying known data\\
    \texttt{Anomaly/Outlier Detection}& Known & Known/Unknown & Identifying rare items\\
    \texttt{Open Set Recognition} & Known & Known/Unknown & Classifying known data and rejecting unknowns \\
    \bottomrule
    \label{tab1}
  \end{tabular}}
\end{table*}

There are many approaches regarding classification with a reject option in the literature \cite{bartlett2008classification, herbei2006classification, fischer2016optimal, yuan2010classification, fumera2000reject, geifman2017selective, wegkamp2007lasso, fumera2002support, grandvalet2009support, zhang2006ro} which have been adjusted to support open sets. In a threshold-based classification strategy, an instance is recognized as unknown if the matching score to the most likely class is below the established threshold, i.e., the sample is far away from all training samples \cite{dubuisson1993statistical, muzzolini1998classifier}. For instance, Phillips et al. \cite{phillips2011evaluation} described an evaluation protocol for open-set face recognition algorithms which decides whether the identity of a sample corresponds to a known class or not if the similarity score exceeds an ad-hoc rejection threshold, and then reports the identity of the accepted sample. Another work \cite{vareto2017towards} estimated a rejection threshold based on the ratio of the two highest decision scores obtained from a vote list ranking. This method combines hashing
functions and classification methods. Whenever a face query is requested, it is compared to all hashing functions and the vote list is generated based on their response values.
A transduction-based study which considered open-set face recognition from an evaluation point of view is introduced in \cite{li2005open}. The proposed open set algorithm uses distances of a test image to its k-nearest neighbors in both inter and intraclass and extracts the credibility values' distribution. This method rejects a face as unknown if the highest credibility value passes a proper threshold. However, defining such a threshold for an unknown is the critical part of all the approaches adopting the threshold-based classification scheme. For instance, in a task of image classification, when an image is slightly different from what the network learned, adding a threshold to the classification output may reject the image as unknown. So, thresholding depends on the operating environment of a recognition system and how distinct the class is. Moreover, this technique does not work very well as all outliers of each class may be classified as unknown and rejected. The other challenge with thresholding is detecting adversarial images trying to bypass machine learning systems to misclassify. \par 
The greatest part of rejection-adapted approaches rested upon variants of Support Vector Machine (SVM) \cite{cortes1995support} classifiers with the ability to reject observations \cite{fumera2002support, zhang2006ro, grandvalet2009support}, and the one-class classifiers based on support vectors  \cite{chen2009hierarchical, hanczar2014combination,scholkopf2001estimating,homenda2014classification,tax2004support}.
Although such techniques are related to OSR in the sense of rejecting an input, they have different reasons to do rejection actions. Classifiers with rejection options focus on the ambiguity between classes to reject an uncertain input of one class as a member of another one and minimize the distribution mismatch between the training and testing domains, while OSR rejects an input because of not belonging to any of the known classes. Chow \cite{chow1970optimum} derived optimal thresholds to optimize the ambiguous regions between classes in multiclass classification task with the assumption of known prior probabilities of classes. Therefore, rejecting uncertain inputs in such classifiers protects misclassification but is not enough to handle unknowns. They have infinite positively labeled open space and infinitely open space risk and thus are not able to solve OSR problems formally. In these techniques, unknowns often appear to be uncertain and are labeled with confidence. In contrast, OSR supports rejecting the unknown object by discovering the acceptable amount of uncertainty and searching among any of the known classes to identify if the true class exists. In other words, the set of possible outcomes of predictions is an important difference between OSR and a typical multi-class classifier.
\par
\begin{figure}[h]
  \centering
  \includegraphics[width=\linewidth]{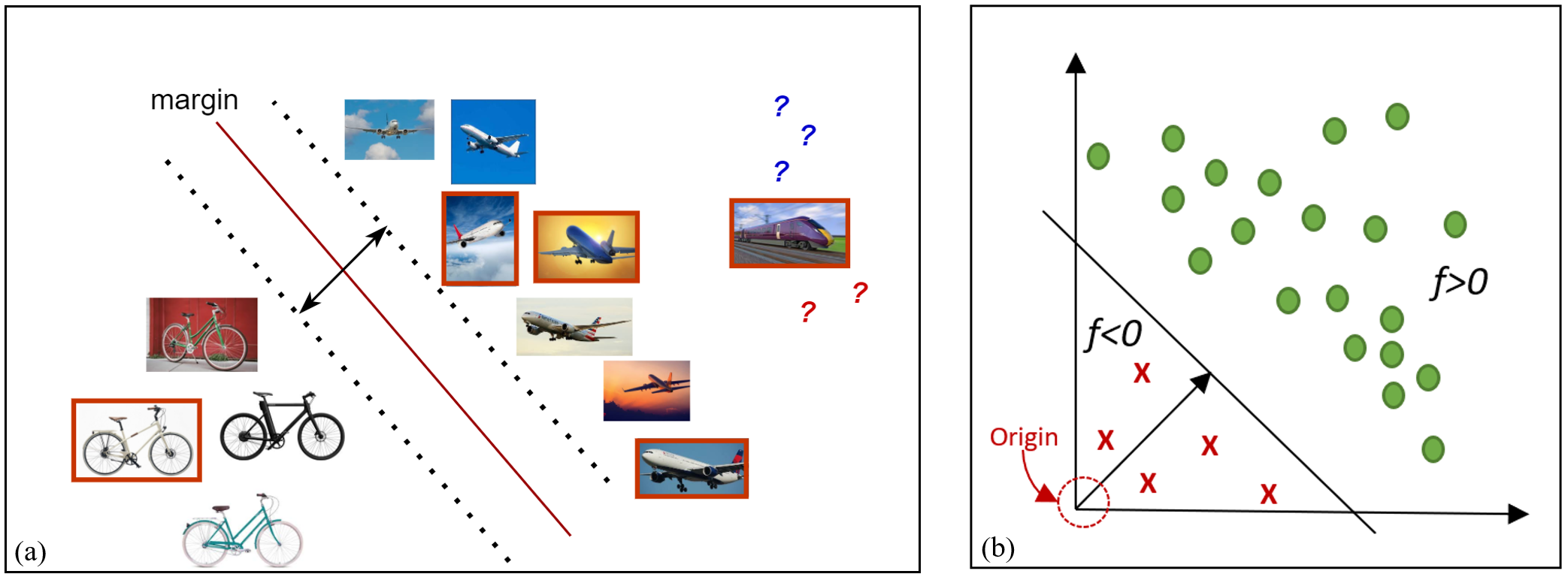}
  \caption{(a) Two-class SVM classifier. Images with an orange box are from testing and the rest of images are from training. Testing images can be known (“bicycle”, “airplane”) or unknown (“train,” “?”). (b) One-class SVM classifier.}
  \label{fig3}
\end{figure}

For example, in margin classifiers like SVMs, confidence is evaluated in terms of an associated distance to the decision boundary given for each example. The goal of SVMs is to find an optimal hyperplane to classify and separate the classes of training samples. The hyperplane defines half-spaces and divides examples of the separate categories by maximizing the distance between itself and the nearest training points. In such classifiers uncertainty is high near the decision boundary and confidence will be increased with distance from the decision boundary; the farther an input is from the margin, the more confident one can be that it belongs to the known classes. Thus, an unknown far from the boundary is incorrectly labeled and will be incorrectly classified with very strong evidence. For example, in {Figure~\textcolor{blue}{\ref{fig3}}(a)}, a plane found by the SVM separates bicycles and airplanes and maximizes the SVM margin making “airplane” a half-space. An unknown (“?”) far from the training data will be misclassified and likely be labeled “airplane” as the label propagation is not limited.


For such classifiers that use observation-to-margin distance as the only information to identify unknowns, resting on a threshold as a confidence rate for rejection is not enough for discovering the hidden unknown classes. Moreover, due to incomplete information about unknown classes, selection of the decision threshold depends merely on the knowledge of known classes, and the decision score calibration is processed implicitly by closed set assumptions. Therefore, OSR cannot use rejection-adapted SVM as a good option, although it outperforms a multi-class SVM which strictly assigns a label to the known. Additionally, there exist limitations in probabilistic models for the open set problem where the prior probability of the classes is unknown and Bayes' theorem is violated. Considering the likelihood of unknown classes, Bayes' rule cannot be exactly utilized as Bayesian posterior probability. This model, which holds closed world assumptions, cannot be modeled for unknown classes unless the probability of all unknown classes is assumed as known. On the other hand, an obvious approach to add a rejection option to a multi-class classifier is to incorporate a thresholded probability model, in which a decision threshold is added into a posterior probability estimator, $P(L\mid X)$ \cite{kwok1999moderating, huang2006generalized}. Where $L \in N$ is a particularly known class label for a fixed set of $N$ known classes and $X$ is an input sample. At the time of the appearance of unknown classes, a given data is labeled as unknown if the maximum probability over known classes is below the defined threshold. However, there is a chance that a misclassification still exists due to unlimited open space risk.
\par
On the other hand, some people argue that identifying novel classes \cite{bishop1994novelty, abati2018and, perera2019learning}, discovering outliers \cite{ritter1997outliers, xia2015learning, you2017provable} and detecting anomalies \cite{chalapathy2017robust, golan2018deep, sabokrou2018adversarially} sometimes can solve the OSR problem \cite{bodesheim2015local, lazzaretti2016novelty, schultheiss2017finding}. Although these methods such as one-class support vector machine (OCSVM) \cite{scholkopf2001estimating} or support vector data description (SVDD) \cite{tax2004support} referred to the problem of identifying unknown data and have been a good start for OSR, the problem setting is different from that of OSR. These techniques are restricted to merely solve OSR for One-class classification problems \cite{tax2001uniform, khan2009survey} in the one-class setting. One-class classification is solved by finding a decision function $f$ which transforms input data into a high dimensional feature space. The function $f$ is positive in some small region corresponding to one class of objects, and negative elsewhere. This algorithm tries to find a separating hyperplane which maximizes the margin between the training data (positive examples) and the origin (considered as negative examples) (see {Figure~\textcolor{blue}{\ref{fig3}}(b)}). An one-class classifier then is trained to label a test example $x$ as an outlier if $f (x) < 0$, and normal if $f (x) > 0$.

Although the possibility of modeling each single class \cite{tax2008growing} or concentrating multiple known classes into a single one opens up a new way for multi-class novelty recognition \cite{bodesheim2013kernel, tang2019lgnd}, these techniques alone are not sufficient for creating a balance between the risks of the unknown and multiclass recognition for OSR, leading to poor performance. Compared to some anomaly detection techniques where an auxiliary dataset of outliers is accessible at the training time \cite{hendrycks2018deep}, OSR problems do not have access to unknown classes. 
A number of surveys have been written to analyze and discuss the concept of outliers/anomalies from different points of view \cite{marsland2003novelty, markou2003novelty, markou2003novelty2, pimentel2014review, chandola2009anomaly, zimek2018there, aggarwal2015outlier, hodge2004survey}. These techniques with a very long history in machine learning can model normal data, then find a distance from the class mean for each sample and place an optimal threshold for discovering abnormalities. In the case of the existence of an appropriate threshold on one or more one-class classifiers \cite{li2005open, tax2008growing, liu2016modular}, a finite open space risk will be produced and OSR can be supported. However, these techniques must have a robust performance that requires trading off between maximizing the recognition rate and minimizing the inclusion of novel data. Moreover, they achieve less stability and worse performance over OSR models once classes are withheld during training.

\begin{table*}[width=1.9\textwidth]
  \caption{Different categories of statistical approaches for OSR}
  \label{tab:commands}
  \resizebox{\textwidth}{!}{%
  \begin{tabular}{ccccl}
    \toprule
    \textbf{Statistical Approaches} & \textbf{Papers} & \textbf{Extreme Value Theory} & \textbf{Exploiting Unlabeled Data} & \textbf{Incremental Learning}\\
    \midrule
    \texttt{\textbf{Rejection-adapted SVM}}& \begin{tabular}{@{}c@{}}[14][107][114][115][120][128] \\ $ $[123][124][125] \\ $ $[16][108] \end{tabular} & \begin{tabular}{@{}c@{}} --- \\--- \\ $\surd$ \end{tabular} & \begin{tabular}{@{}c@{}} --- \\ $\surd$ \\ ---  \end{tabular} &
    \begin{tabular}{@{}c@{}} --- \\ --- \\ ---  \end{tabular} \\
    \texttt{ } &  &  &  & \\
    \texttt{\textbf{Nearest Class Mean}}& \begin{tabular}{@{}c@{}}[3][118][121] \\ $ $[133] \end{tabular} & \begin{tabular}{@{}c@{}} --- \\---  \end{tabular} & \begin{tabular}{@{}c@{}} --- \\ ---   \end{tabular} &
    \begin{tabular}{@{}c@{}} $\surd$ \\ ---  \end{tabular} \\
    \texttt{ } &  &  &  & \\
    \texttt{\textbf{Sparse Representation}} & [130] & $\surd$ & --- & ---\\
    \texttt{ } &  &  &  & \\
    \texttt{\textbf{Margin Distribution}} & 
    \begin{tabular}{@{}c@{}}[111] \\ $ $[131] \end{tabular} & \begin{tabular}{@{}c@{}} $\surd$ \\$\surd$ \end{tabular} & \begin{tabular}{@{}c@{}} --- \\ ---  \end{tabular} &
    \begin{tabular}{@{}c@{}} --- \\ $\surd$   \end{tabular} \\
    \bottomrule
    \label{tab2}
  \end{tabular}}
\end{table*}

\section{A Categorization of OSR Techniques}
We summarize statistical approaches into four categories. {Table~\textcolor{blue}{\ref{tab2}}} shows these four categories and the related papers to each one. Besides, we represent if the corresponding methods adopt Extreme Value Theory, Exploiting Unlabeled Data, and Incremental Learning. {Table~\textcolor{blue}{\ref{tab3}}} also represents existing OSR methods based on Deep Neural Networks and shows if any of these methods adopt Extreme Value Theory.

\subsection{Statistical approaches}
Scheirer et al. \cite{scheirer2012toward} introduced the first formalization of OSR by balancing open space risk $R_{O}$ associated with labeling data that is far from known training samples against minimizing empirical risk $R_{\epsilon}$ over training data. By assuming $f$ to be a measurable recognition function, open space risk $R_{O}(f)$ for known class $k$ is described as the following: 

\begin{equation}
R_{O}(f)= \frac{\int_{O}^{ }f_{k}(x)dx}{\int_{S_{k}}^{ }f_{k}(x)dx}
\label{eqn:1}
\end{equation}
This formalization provides the proportional value of positively labeled open space $O$ against to the total measure of $S_{k}$ consisting all of the known positive training samples $x\in k $ as well as the positively labeled open space. This paper argued that the essential element of OSR is finding the recognition function $f$, where $f(x) > 0$ indicates the positive recognition of the class $k$ of interest. This function is defined as a minimization of the open space risk to capture the risk of labeling the unknown samples as known, beyond the sensible recognition of the training data, as follows.
\begin{equation}
argmin_{f \in H} \left \{ R_{O}(f)+\lambda_{r}R_{\epsilon}(f)\right \}  
\end{equation}
where $\lambda_{r}$ is the regularization tradeoff between open space risk and empirical risk. 
This study proposed a “1-vs-set Machine” which consists of two parallel hyperplanes. The proposed formulation with a linear kernel balances empirical and open space risk by exploiting a second hyperplane from the marginal distances of a 1-class or binary SVM. The main hyperplane is a base SVM which defines half-spaces and aims at maximizing the margin. The second hyperplane is added in such a way as to minimize the positive labeled region bounded between two planes and handle open space risk. This method defines a definition that generally describes region of known classes for each individual binary SVM, however, it lacks the procedure of distance measurements. It even does not clarify the space for measurements of such distances. Therefore, it cannot bound the space that each known class belongs to that leads to the existence of open space risk. Inspired by this technique, Cevikalp \cite{cevikalp2016best} found the best fitting hyperplanes by placing them close to the samples of one class and far from the other class samples.

\begin{table}
  \caption{Different categories of DNN-based approaches for OSR}
  \label{tab:commands}
  \resizebox{\columnwidth}{!}{%
  \begin{tabular}{ccccl}
    \toprule
    \textbf{ DNN-based Approaches} & \textbf{Papers} & \textbf{Extreme Value Theory} \\
    \midrule
    
    \texttt{\textbf{Adversarial Learning}}& \begin{tabular}{@{}c@{}}[15] \\ $ $[162-166] \end{tabular} & \begin{tabular}{@{}c@{}} $\surd$ \\---  \end{tabular} \\
    \texttt{ } &  &  &  & \\
    \texttt{\textbf{Background-class-based Modeling}} & [139] [153-155] & --- \\
    \texttt{ } &  &  &  & \\
    \texttt{\textbf{Others}} & 
    \begin{tabular}{@{}c@{}}[2][110][138][140][167][174] \\ $ $[39][134-137][152][175][176] \end{tabular} & \begin{tabular}{@{}c@{}} $\surd$ \\ --- \end{tabular} \\
    \bottomrule
    \label{tab3}
  \end{tabular}}
\end{table}
\par
There have been many more attempts over the past years to address open space risk for training OSR models. Followup works by Scheirer et al. \cite{scheirer2014probability, jain2014multi} were inspired by the fact that leveraging Extreme Value Theory (EVT) \cite{castillo2012extreme} on the SVM decision scores provides better performance than exactly applying the raw score values. Both approaches have proposed EVT-based SVM calibration techniques to enable the SVM-based classification to deal with an open-set setting. Other OSR algorithms such as \cite{bendale2016towards, yoshihashi2019classification, rudd2017extreme, scheirer2017extreme} also include EVT to analyze the association of a data point with an unknown class. Extreme value modeling has been increasingly used to analyze post-processing scores and enhance the performance of OSR. This theory is meant to study a level of confidence by determining the fraction of objects deviating from the expected value.
EVT is effectively used in many research areas such as environmental risk management, finance, insurance, anomaly detection, or network monitoring. 
The following is the definition of EVT:\par
Let $\{X_{1},X_{2},X_{3},...,X_{n}\}$ be a sequence of independent random variables with unknown distribution function $F(x)$, and $X_{m}= \underset{i}{max}X_{i}$ ,  $i\in[1,n]$. Assume there exists a pair of sequences $(a_{n},b_{n})$ with $a_{n} > 0$ and $b_{n} \in \mathbb{R}$ such that $\lim_{n\rightarrow\infty} P\left (\frac{X_{m} - b_{n}}{a_{n}}  \right ) = G \left ( x \right )$. Then if $G$ is a non-degenerate distribution function, it belongs to one of Fréchet, Weibull, or Gumbel distribution families. These three distributions can be combined into a single general form which is called  Generalized Extreme Value (GEV) distribution and is defined in Equation (\textcolor{blue}{\ref{3}}):
\begin{equation}\label{3}
E\left ( x;\mu,\sigma,\xi \right ) = \exp\left \{  -\left [ 1+\xi \left ( \frac{x-\mu }{\sigma } \right ) \right ]^{\frac{-1}{\xi }}  \right \}
\end{equation}
Where $\mu$, $\sigma$ and $\xi$ are the location, scaling, and shape parameters, respectively. These three models provide the data distribution for a reasonable evaluation of the probability of occurrence of rare events. As extreme values appear in the tails of the distributions, EVT examines the distribution tails and aims to predict the probability that a given sample is an extreme value applying Equation (\textcolor{blue}{\ref{3}}). For OSR, EVT models the probability distributions of the match and non-match recognition scores and the rejection threshold is usually estimated from the overlap region of extreme values found in the tails of probability distributions. 
\par
Unlike SVM, which divides all the space with hyperplanes to define sections and allocate them to one of the current classes, Scheirer et al. \cite{scheirer2014probability} applied EVT and Weibull distributions to build such hyperplanes without dividing the whole space. This work introduced the idea of a compact abating probability (CAP) model based on a one-class classifier which, when thresholded, can further limit the open space risk. The labeled region is limited and the open space risk is minimized if the value of the probability of class membership is decreasing in all directions as samples pull out of the training data and move towards the open space. Distribution of decision scores for unknown recognition is considered by extending “1-vs-Set Machine” to W-SVM (a Weibull-calibrated non-linear classifier). This algorithm yields better modeling for a binary SVM at the decision boundaries by applying EVT for score calibration and combining One-Class SVMs using a Radial Basis Function (RBF) kernel with the scores from multi-class SVM. The first application of the W-SVM algorithm is to the difficult problem of fingerprint spoof detection where inter-class distances between a live finger and an effective spoof one are small in the feature space \cite{rattani2015open}. Another work \cite{cruz2017open} based on W-SVM is proposed for open set intrusion detection on the KDDCUP'99 dataset. 
\par 
Based on this intuition, Jain et al. \cite{jain2014multi} introduced a variant of W-SVM that is called Support Vector Machines with Probability of Inclusion (PISVM). This algorithm formulates the multi-class OSR problem as one of the modeling positive training data at the decision boundary. An SVM with RBF kernel is utilized as a binary classifier for each class and trained by the One-vs-All approach, where the samples of the remaining classes are assumed as negative. It models the unnormalized posterior probability of inclusion for multiple classes as a basis to reject unknown samples. Then, it fits probability distributions consistent with the statistical EVT, leveraged on the decision scores from the positive training samples. For a given sample, a class is chosen whose decision value makes the maximum probability of induction. The sample is recognized as unknown if that maximum is under a predefined threshold. Although the proposed algorithm is more accurate than W-SVM, it did not always confine open space risk, the issue that occurred with a regular SVM.
\par
In spite of being a recent research focus, EVT for OSR does not merely guarantee a bounded open space, as PISVM \cite{jain2014multi} does not always bound open space and W-SVM \cite{scheirer2014probability} relies on one-class models to bound open space rather than counting on EVT models. Compared to EVT-based models, a simpler algorithm called Specialized SVM was proposed by Júnioret et al. \cite{junior2016specialized} recently. This algorithm bounds the represented space for known categories and provides a finite risk of the unknown if using an RBF kernel and limiting the bias term to be negative. Additionally, the proposed EVT-based calibration of 1-vs-rest RBF SVMs modeling in both W-SVM and PI-SVM has two deficiencies. The first deficiency is that it is not ideal for OSR which requires incremental updates. Supporting incremental learning is a principal goal in designing an algorithm for OSR, especially one that is constantly used over a long period of time. This approach cannot add novel detected objects and tune the model to enhance the fit as a new item arrives. So it is not able to learn the model incrementally. The second deficiency is that it does not address the fundamental issue of choosing thresholds, which requires prior knowledge in such threshold-based classification models. However, the authors of the last two papers recommended choosing the thresholds according to the problem openness, which is not reasonable since the openness is not usually known in the corresponding problem.
 \par
To tackle these deficiencies Scherreik et al. \cite{scherreik2016open} formulated the probabilistic open space SVM (POS-SVM) which rests on a one-vs-all binary SVM. An individual reject threshold for each of the known classes is computed and optimized by a validation set. Platt's method \cite{platt1999probabilistic}, the most widely used probability estimator, is also used to convert SVM scores to a calibrated probability estimation.
Another possible approach more appropriate for the open-world with incremental learning capabilities proposed a Nearest Non-Outlier (NNO) algorithm \cite{bendale2015towards}. NNO adapts the Nearest Class Mean Classifier (NCM) \cite{mensink2013distance}, the basis of most open-set classifiers, for OSR by using non-negative combinations of abating distance. This work is built on the concept of a CAP model; however, it generalizes the model to gain zero open space risk by applying a threshold on any non-negative combination of abating functions. NCM represents the classes by the mean feature vector of their components, and a test sample is set to a class with the closest mean using Euclidean distance between the class mean and the test feature vectors. The NNO algorithm was inaccurate because of using thresholded distances from the nearest class mean. Additionally, it does not tune the rejection threshold automatically as new classes arrive and the problem evolves. So this algorithm does not properly model the dynamic nature of open-world recognition. To mitigate this problem, Rosa et al. \cite{de2016online} used the Hoeffding bound \cite{hoeffding1994probability} to incrementally update the threshold for an unknown class, instead of estimating it from an initial set of known classes and keeping it fixed as previously used in \cite{bendale2015towards}. 

\par
Statistical approaches such as threshold-based decision technique are being widely employed in text document open-set classifications \cite{fei2016breaking,doan2017overcoming}.
Probably, cbsSVM \cite{fei2016breaking} is the first open multiclass text classifier. This model is based on the CBS (Center-Based Similarity) space learning method \cite{fei2015social}, whereby a center for each class in the original problem is computed first. Then the data is transformed into a vector of their similarities to the class centroids to limit positive labeled area from an infinite space to a finite space.
A decision threshold is then applied on posterior probabilities which are estimated from the SVM scores for each classifier using Platt's algorithm \cite{platt1999probabilistic} to identify unknown classes.
\par
Doan et al. \cite{doan2017overcoming} represented Nearest Centroid Class (NCC) which is incremental learning and built upon the NCM algorithm. Instead of using the class mean for each class member, this model is based on a series of closest neighbors of the centroid class. In spite of its similarity with NNO in terms of using multiple centroids, the proposed model addresses the issue of the new classes being added incrementally related to NNO and updates information for a class ball. During training, this algorithm attempts to create the boundary region for each known class. Each class is a set of balls centered at class centroids where each ball represents a number of its data points. An observation is treated as an unknown when not any of the nearest class boundaries support it. \par 
Additional works like Assign-and-Transform-Iteratively (ATI) \cite{panareda2017open}, LACU (Learning with Augmented Class with Unlabeled data) framework \cite{da2014learning} and Separate to Adapt (STA) \cite{liu2019separate} require the help of unknown source samples. Additionally, these methods maintain the assumption of containing unknown classes in the source domain. Busto et al. \cite{panareda2017open} utilized unknown source samples whose class does not overlap with that of the unknown target. This algorithm maps the source domain's feature space to the target domain. It learns this association by minimizing the distance from target samples to each of the source classes' center. Based on a binary linear program, the assignment problem is defined that also implicitly handles outliers by discarding predicted unknown target samples not connected to any of the source domain's samples. This process iterates over the converted source samples to repeat the process of solving the assignment problem, approximating the mapping from one domain to another one, and updating the transformation until it converges. After convergence, linear SVMs are trained in a one-vs-one setting over the converted data to label the target domain. In this work, the execution of a typical SVM is compared with an alternative model introduced in \cite{scheirer2014probability}. 
The other work \cite{da2014learning} presented the LACU-SVM approach to address OSR by exploiting an unlabeled dataset besides the training set and tuning the decision boundary. Based on the large margin principle from the SVM algorithm, classes should be divided by large margin separators. Thus, the unlabeled data can identify large margin separators that have similar performance to the seen classes when adopting the one-vs-rest approach. LACU then selects one of these separators that is closest to the labeled region. Distinguishing augmented (unknown) classes involves the utilization of the LACU-SVM in which seen classes are surrounded by large margin separators. Then, it picks a classification boundary among all low-density separators that minimizes the misclassification risks among the seen classes as well as between the augmented and the seen classes simultaneously.
\par
Unlike several methods proposed in the literature to address OSR, Liu et al. \cite{liu2019separate} recently took into account the openness \cite{scheirer2012toward} of the target domain, which is measured by the proportion of unknown classes to be identified in the target domain. In this work, a multi-binary classifier is trained in a one-vs-rest setting to measure the similarity between the entire target domain and each source class. All the target samples are ranked by such similarity. Then, a binary classifier is trained using samples with the highest/lowest similarity to separate all target samples and generate the weights for rejection. These two steps are repeated and samples of unknown classes are rejected progressively in the adversarial domain where one more class is added to the source classifier for the unknown class.
\par
Web genre identification (WGI) is considered as a multi-class text classification task with the ability to automatically recognize the genre of web documents. Therefore, search results can be categorized based on the genres that not only facilitate retrieving information but also provides rich descriptions of documents and enables more specialized queries. Instead of the content, WGI puts the emphasis on the relation of form and style with their associated web pages \cite{mason2009n, rosso2008user}.
In an experimental study on the open-set classification models for WGI setup, Pritsos et al. \cite{pritsos2018open} examined one-class SVMs and Random Feature Subspacing Ensembles (RFSE) \cite{pritsos2013open} models.
With respect to this fact that most of the complementary information to differentiate known from unknown samples is placed in the tail of a distribution, modeling the tail of match and non-match error distributions can help to find the optimal threshold for a given recognition model. Inspired by this intuition, Zhang et al. \cite{zhang2016sparse} extended the Sparse Representation-based Classification (SRC) algorithm to OSR. This algorithm models the tails of these two residual errors using EVT. The identity of an unknown test sample and open-set identification is determined by getting the confidence score for that sample and hypothesis testing.\par
Extreme Value Machine (EVM) \cite{rudd2017extreme} as a probabilistic framework for open set classification also considers Weibull distributional information when learning recognition functions. EVM is the first classifier to perform a nonlinear RBF approach motivated by EVT and provides a more powerful representation model for OpenMax which will be discussed in section \ref{section3.2}. Using CAP models, EVM is able to bound open space. \cite{henrydoss2017incremental, gunther2017toward} are applications of EVM in intrusion detection and open face recognition respectively.  However, this approach has drawbacks with regard to the choice of the threshold which controls the open set classification error and more important, strongly relies on the relative arrangement of the known classes. EVM assumes that the behavior of the unknowns can be inferred by the geometry of the known classes, and thus the recognition task may fail when the known and unknown geometries of classes are different.
To overcome these limitations, two robust algorithms \cite{vignotto2018extreme} derived from EVT that do not rely on the geometry of the observed data. These classifiers, called generalized Pareto distribution (GPD) and generalized extreme value (GEV), utilize the intuition that new points to be classified as known or unknown are more likely to be unknown if they are far away from the training data. Moreover, these algorithms are efficient to update upon arising new training data.
\par
There are also a few studies \cite{pritsos2019open,junior2016specialized,juefei2016multi} utilizing Nearest Neighbor models on this topic. Júnior et al. 
\cite{junior2017nearest} proposed the Nearest Neighbor Distance Ratio (NNDR) classifier, which in turn, is a multiclass open-set extension for the Nearest Neighbor (NN) algorithm and is referred to as Open Set NN (OSNN). 
During the prediction phase, the OSNN first finds the nearest and second nearest neighbors $y$ and $z$ regarding a test sample $t$ in order that $\omega(y)\neq \omega(z)$, where $\omega(s) \in L=\left \{ l_{1},l_{2},...,l_{n} \right \}$ represents the class of sample $s$ and $L$ is a set of training labels. Then, this classifier calculates the similarity scores' ratio and applies a threshold to recognize sample $t$ as unknown having low similarity. This ratio is defined by $r=d(t,y)/d(t,z)$, where Euclidean distance of two samples $s$ and $s^{'}$ is shown by $d(s, s^{'})$. Recently, Pritsos et al. \cite{pritsos2019open} viewed WGI as an open-set task and applied the NNDR algorithm to its setup to better deal with incomplete genre palettes.  

\subsection{Deep neural network-based algorithms} \label{section3.2}
\begin{figure}[h]
    \centering
    \includegraphics[width=\linewidth]{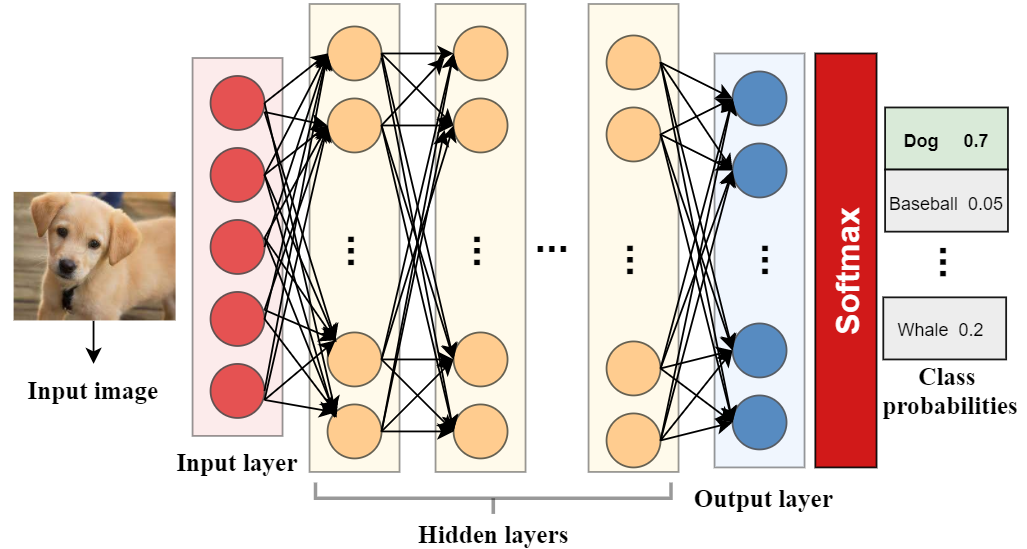}
    \caption{Schematic structure of DNNs. The example of an image classification task which takes in an image and outputs the confidence scores for a predefined set of classes.}
    \label{fig4}
\end{figure}

An Artificial neural network inspired by the human brain simulates the organization and learning of biological neurons. It is composed of the neurons or processing-computing units which are interconnected to each other and organized in three types of layers called input, hidden, and output layers. The input layer receives data and communicates to the hidden layer(s) where the actual processing is done using the weighted paths. Then, the hidden layers connect to the last layer in the network to give the output. Additional hidden layers provide more flexibility to the network and make it more powerful to model complex relationships between input and output. Such a neural network with many hidden layers is called Deep Neural Networks (DNNs) (see {Figure~\textcolor{blue}{\ref{fig4}}}). \par

Neurons represented by the circles communicate with each other by sending signals over several weighted connections. Weights are basically the effect of previous layers’ neurons on the ones of the current layer. Every neuron has a state of activation that is the output of the neuron and goes to the next layer. The activation values of neurons in one layer act as the input for the activation function of the next layer, where the weight connections define the amount of this contribution. This is how the activation values of the current layer are updated in the forward propagation process. In a multi-class problem, typically, most of the DNNs use the Softmax function as the activation function in the output layer. 
The network is trained in the backward propagation by adjusting the value of weighted connections. The objective of training is to minimize the loss function, such as cross-entropy that represents the difference between the output of the softmax function and the desired output to achieve a low classification error in the training data. This process leads to a set of properly adjusted weights that enables the neural network to be used effectively for the purpose it is initially designed for. Finally, during the testing time, the Softmax function represents the probability that the sample $x$ is labeled with class $n$. Due to its closed nature, a deep network links an unknown sample to the class with the maximum score given by Softmax, leading to misclassification of that sample.

\par
\begin{figure}[h]
    \centering
    \includegraphics[width=\linewidth]{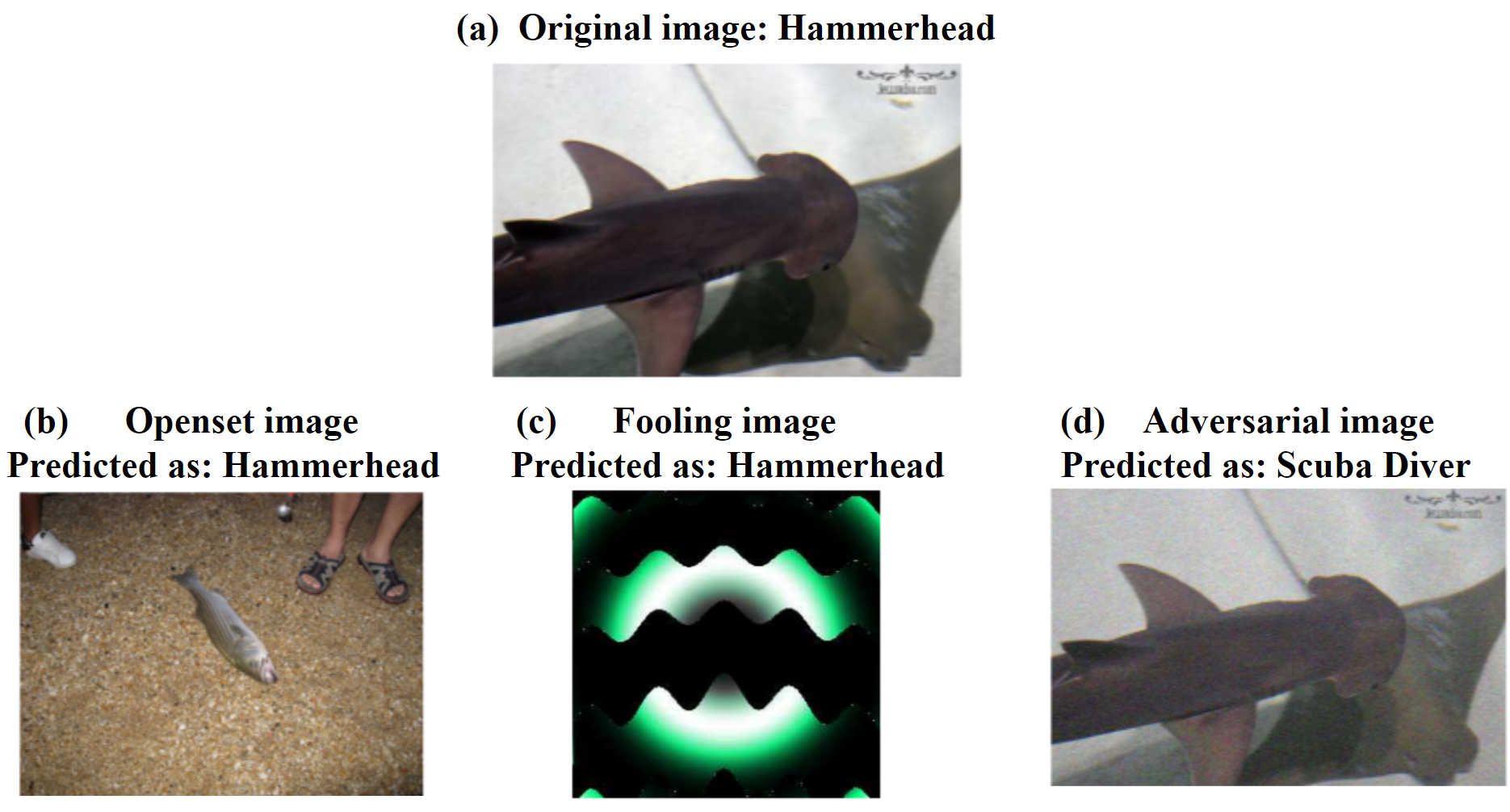}
    \caption{Examples of an original, open set, fooling, and adversarial images taken from \cite{bendale2016towards}. (b) Example of a real image from an unknown category which is mapped to the class with the maximum response provided by Softmax. (c) A fooling input image which is unrecognizable to humans, but DNNs believe with high certainty to be a Hammerhead. (d) An adversarial image specifically constructed from hammerhead to scuba to fool DNNs into making an incorrect detection.}
    \label{fig5}
\end{figure}

Following the extensions of traditional classification algorithms for OSR, there is a considerable amount of research in developing deep neural networks for OSR in the literature \cite{bendale2016towards, hassen2018learning, ge2017generative, cardoso2015bounded, cardoso2017weightless, yoshihashi2019classification, shu2017doc, shu2018unseen, oza2019c2ae, dhamija2018reducing, rozsa2017adversarial, venkataram2018open}. 
However, with the shift to deep networks, which combines learning features and learning the classifier, the performance of the system for OSR is still far from optimal \cite{boult2019learning}.
Researches have addressed this problem by thresholding on the Softmax scores. It seems that for an unknown sample this function produces low probability for all the classes so that thresholding on the output probability can help to reject the unknowns. However, combining a deep network with thresholded probabilities determines uncertain predictions which are a small part of unknown inputs. As a consequence, thresholding Softmax is not enough to detect fooling or adversarial examples. Researches such as \cite{nguyen2015deep, goodfellow2014explaining} have shown that DNNs are particularly vulnerable to these examples and are easily fooled. Fooling examples target the desired class and seek to increase the corresponding probability of that class. These artificially constructed examples are fully imperceptible to humans, but the classifier sees them as a member of the desired classes and labels them with high certainty (see {Figure~\textcolor{blue}{\ref{fig5}}(c)}). A more restrictive case is rejecting an adversarial example \cite{szegedy2013intriguing}-- a visually similar input to the training dataset with small but intentional perturbations, such that it is mislabeled by a classifier as an entirely different class with high confidence (see {Figure~\textcolor{blue}{\ref{fig5}}(d)}).

\par 
The difficulty level of rejecting adversarial examples depends on how close the example is to the target class. For instance, if an adversarial example like a salmon shark is produced from a nearby class like a hammerhead, it will fail to be rejected as an unknown. However, if this example is generated from a faraway target class like scuba, it will be rejected as an unknown due to a remarkable difference in the output scores. That is why most of the studies proposed for OSR do not consider these examples in their experiments. However, there are many defenses proposed in the literature to enhance the robustness of a neural network against adversarial examples \cite{yuan2019adversarial, papernot2016distillation, liang2017detecting, feinman2017detecting, grosse2017statistical}. Ten detection strategies are evaluated in \cite {carlini2017adversarial} to demonstrate how these defenses can be evaded. To mitigate the fooling problem in neural networks, the space around training data is tightened in \cite{kardan2017mitigating} that makes the secure classification of examples only within that space. For the sake of diversity, several outputs are assigned to each class and compelled to compete. In other words, in the same class, they secure their input space among output units that guarantees the proximity of outputs to the training instances.
\par
Moreover, a more effective rejection solution than thresholding softmax is using a garbage or background class which has dominated most of the modern detection approaches like \cite{ren2015faster, zhang2017towards, liu2016ssd}. Such background-class-based modeling can tackle the problem of unknowns in neural networks by adding another class as representative of unknown samples during training. Although this approach works well for datasets like PASCAL \cite{everingham2010pascal} and MS-COCO \cite{lin2014microsoft}, it is a probable source of negative dataset bias \cite{tommasi2017deeper} and has limitations in the real world with infinite negative space of infinitely many unknown inputs to be rejected. Recently, Dhamija et al. \cite{dhamija2018reducing} combined SoftMax with the Entropic Open-Set and Objectosphere losses considering the background and unknown training samples. These losses increase SoftMax entropy for unknown inputs while minimizing the Euclidean length of deep representations of unknown samples. This modification increases separation in deep feature space and improves the handling of background and unknown classes.
\par
\begin{figure}[ht]
    \centering
       \includegraphics[width=\linewidth]{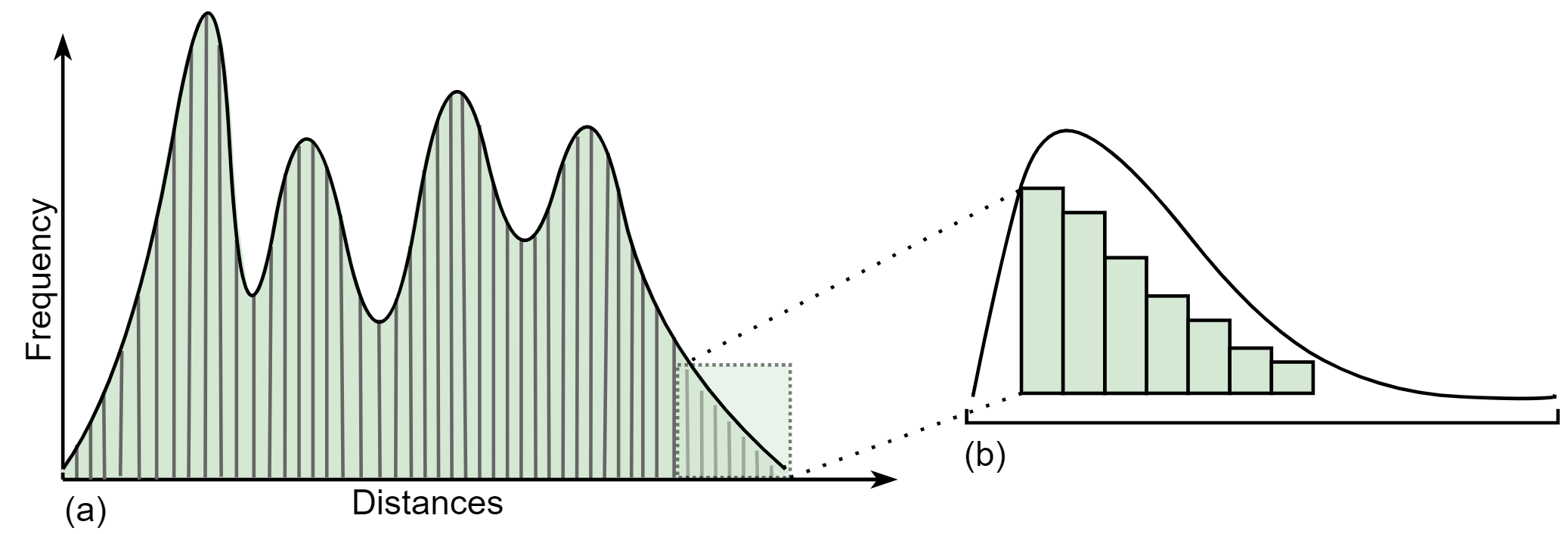}
    \vspace{-0.7em}
    \caption{(a) Class particular distance distribution based on measured distance between each correctly classified training example and the associated MAV. (b) A Weibull distribution related to a certain number of the largest such distances which is fitted separately for each class.}
    \label{fig6}
\end{figure}

The OpenMax \cite{bendale2016towards} proposed by Bendale et al. in 2016 was the first deep open-set classifier without using background samples. Since then, few deep open-set classifiers have been reported. OpenMax does not directly focus on the recognition of adversarial inputs, although it supports the rejection of fooling and unknown images. Rozsa et al. \cite{rozsa2017adversarial} compared DNNs using the traditional Softmax layer with Openmax on their robustness to adversarial examples. Although Openmax is more robust than Softmax to adversarial examples and outperforms networks with thresholding SoftMax, it does not provide robustness to sophisticated adversarial construction techniques. This work adapts the concept of Meta-Recognition \cite{scheirer2011meta} on activation vectors to formally solve OSR for image classification. OpenMax uses the EVT model built from the positive training samples to define a per-class CAP model that can bound the risk of open space to reject unknown inputs by thresholding the model. First, activation vector of each training instance $X$ is computed and defined as $V(X) = \nu_{1}(x),...,\nu_{N}(x)$ for each class, $y=1,...,N$. Mean Activation Vector (MAV) is also computed for each class separately over only correctly classified training examples utilizing NCM concept \cite{mensink2013distance, ristin2014incremental}. Then a Weibull distribution will be fitted to each class on the largest distances between the MAV class and positive training instances (see {Figure~\textcolor{blue}{\ref{fig6}}}). Then, parameters of Weibull distribution for each class are estimated. Let $\rho_{j}=(\tau_{j},\lambda_{j},\kappa_{j})$ be an estimation of EVT Meta-Recognition model for class $j$. Where $\tau$, $\lambda$ and $\kappa$ are the location, shape and scale parameters of the Weibull distribution. After pretraining of a deep neural network and computing the activation vectors and the per-class mean vector, the weights for the $\alpha$ largest activation classes are computed to scale the Weibull probability as the following:
\begin{equation}
\omega_{i}(x) = 1-\frac{\alpha-i}{\alpha } e^{-\left ( \frac{\left \| x-\tau _{(i)} \right \|}{\lambda_{(i)} } \right )^{k_{(i)}}}   i=1,...,\alpha
\end{equation}
During testing, the revised OpenMax activations which are calibrated based on the probabilities from the Weibull distribution are computed, including the unknown class (Equation (\textcolor{blue}{\ref{5}})). Finally, the Softmax layer is used to calculate and update class probabilities on the new activation vectors' values (Equation (\textcolor{blue}{\ref{6}})). 
\begin{equation} \label{5}
\hat{V}\left ( x \right )=V\left ( X \right )\omega(X)
\end{equation}

\begin{equation}\label{6}
\hat{P}\left ( y=j\mid X \right )=\frac{e^{\hat{V}_{j}\left ( x \right )}}{\sum_{i=0}^{N}e^{\hat{V}_{i}\left ( x \right )}}
\end{equation}
The OpenMax approach determines unknown inputs when the probability of the unknown class which is defined to be at index 0 in this paper, has the largest value. This maximum probability
is then subject to the uncertainty threshold to support the rejection of uncertain inputs as well.

OpenMax does not enhance the feature representation for better unknown detection. The challenge of using the distance from the MAV is that the class instances are not projected directly around the MAV by using normal loss functions, such as cross entropy. Moreover, because the testing distance function is not used during training, it might not necessarily be the right distance function for that space. 
\par
\begin{figure}[h]
    \centering
       \includegraphics[width=\linewidth]{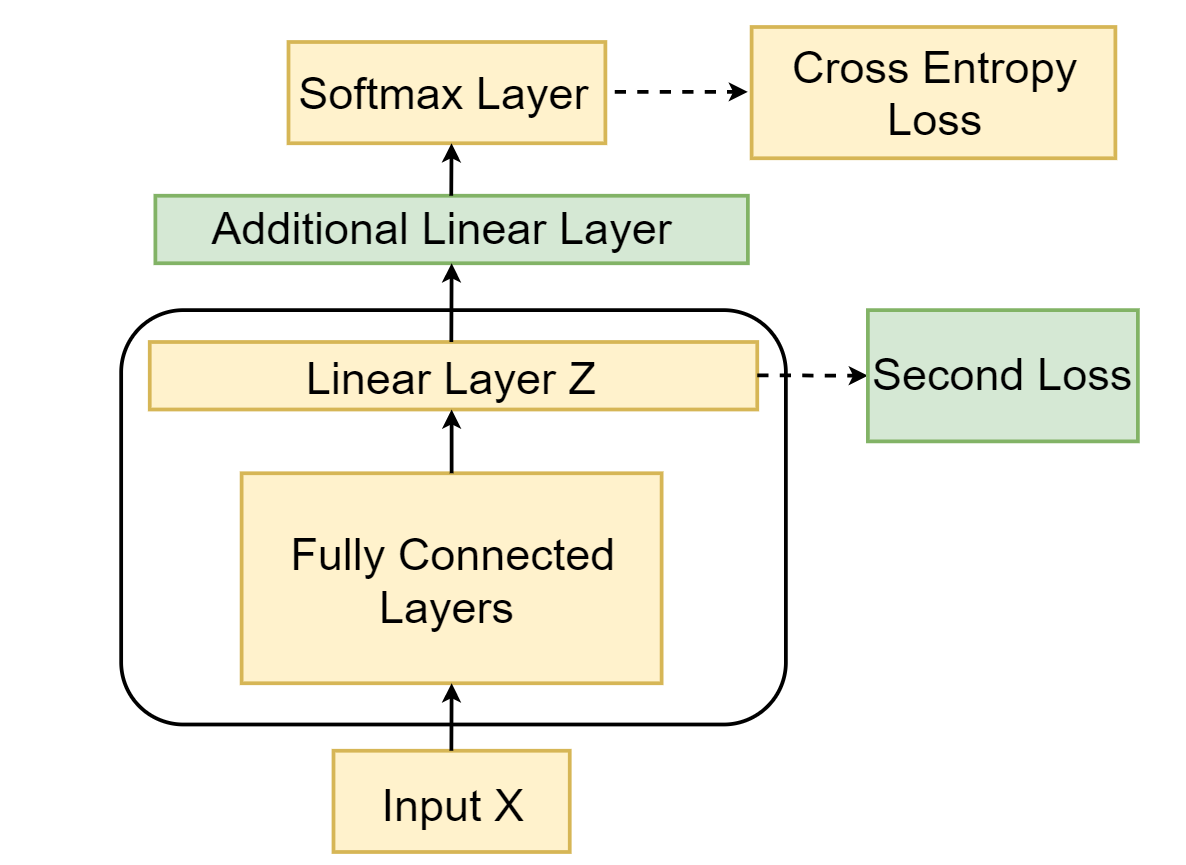}
    \vspace{-0.7em}
    \caption{Combining Cross Entropy Loss with the Second Loss.}
    \label{fig7}
\end{figure}

To address this limitation, Hassen et al. \cite{hassen2018learning} proposed a new and more effective instance representation. Considering $x$ as an instance, the hidden layers in a neural network can be defined as different representations of $x$. This representation can be learned in such a way that instances of the same class are closer together, and instances of different classes are further apart. As a result, unknown classes can be filled in larger spaces between known classes and can be detected more effectively. 
In contrast to a typical neural network in which the output vector $z$ of the final linear layer is served as input to a Softmax layer, in this new setting, the output vector is considered as the projection of the input vector $x$ to a different space. So, in the process of learning, the network is trained to minimize a second loss function. This loss function tries to maximize distances between different classes and minimize the distance of an input from its class mean. Then the distribution of known classes is produced by passing the output of an additional linear layer through the Softmax function (see {Figure~\textcolor{blue}{\ref{fig7}}}). Finally, the network is trained on both cross entropy and a second loss to obtain the low misclassification error in the training set. During testing, a defined threshold on the outlier score identifies known class instances from unknown class instances. The distance between an instance and the closest class mean indicates the outlier score which is the degree to which the network predicts an instance to be an outlier.

\par
Recently, there exist also abundant research on OSR based on the scheme of Generative Adversarial Networks (GANs) \cite{goodfellow2014generative}. A GAN which recently stands out among various deep neural networks consists of a generator and a discriminator. Generally, the generator produces synthetic samples and the discriminator learns to decide if a sample is obtained from the generator or the real dataset. G-OpenMax \cite{ge2017generative} extends OpenMax in adversarial settings and applies GANs to generate unknown instances. These synthetic instances are utilized as an extra training label apart from known labels to adjust the classifier and estimate the probability of unknown classes. The proposed data augmentation technique which is applied to two datasets of hand-written digits and characters has shown itself to be an enhancement of the unknown class identification. However, using it over natural images does not show any performance improvement due to the difficulty of generating plausible images with respect to the training classes as candidates to represent unknown classes.
\par
Along with a similar motivation, another GAN-based approach which is more effective than G-OpenMax for OSR was proposed by Neal et al. \cite{neal2018open}. This strategy, which is called counterfactual image generation, searches for synthetic images by adopting an encoder-decoder GAN technique. These images, referred to as counterfactual-images, are a member of unknown classes but they look like known classes. Using the GAN framework, another work \cite{jo2018open} aims at generating synthesis data which were served as fake unknown classes for the classifier to make it robust against real unknown classes. Yu et al. \cite{yu2017open} proposed the adversarial sample generation (ASG) framework that produces unseen class data. Besides neural networks, ASG can be applied to several learning modes. Inspired by GANs, Yang et al. \cite{yang2019open} proposed a novel model called Open-GAN. In this model, fake target samples are constructed from the generator automatically. Afterwards, the discriminator is modified to adapt multiple classes together with an unknown class. 
Another deep learning method \cite{saito2018open}, Open Set Back Propagation (OSBP), utilizes adversarial training for a more challenging open-set framework which does not require unknown source samples. This approach trains a feature generator to extract features that distinguish known target samples from unknown. Training this generator moves target samples away from the boundary and leads the probability of an unknown target sample to deviate from the pre-defined threshold. These features are then taken by a classifier to place a separator between source and target data and output the probability of target samples to reject unknowns. 
\par
There is also recent interest in exploiting deep neural networks for applying OSR to text classification (\cite{prakhya2017open, prakhya2017open2, grave2017unbounded, shu2017doc, shu2018unseen, cvitkovic2018open, venkataram2018open}). 
Inspired by OpenMax, Prakhya et al. \cite{prakhya2017open} developed an incremental convolutional neural network (CNN)-based text classifier. In contrast to OpenMax, which applies a single mean activation vector, this approach finds the $k$ medoids of every trained class. Compared to image classification, they believe this method represents a class more accurately due to a much smaller number of classes. Then, the distances between the class activation vectors and the corresponding $k$ class medoids are calculated. Applying the average of the $k$ distances, a Weibull model is made for every training class that returns a probability of inclusion of the respective class. Open-set probability is then defined by subtracting the sum of all inclusion probabilities (total closed-set probability) from 1. A sample is either labeled as unknown, if the total open-set probability exceeds the maximum closed-set value, or assigned the class with the highest closed-set probability. Another work \cite{venkataram2018open} combined CNN classification and three outlier detection methods to analyze the output vector of CNN and identify an unknown class. 
\par
A DOC (deep open classifier) proposed by Shu et al. \cite{shu2017doc} is a variant of the CNN \cite{kim2014convolutional} architecture for text classification. This method was compared with OpenMax and represents better performance. One issue related to OpenMax is classifying samples which are difficult to handle as these samples are mostly classified as members of unknown classes. The proposed classifier addresses this issue where the SoftMax layer is replaced by a one-vs-rest final layer of sigmoid activations. The network is trained using a novel loss function to perform joint classification and unknown detection and reduce the open space risk. They show that the risk of open space is reduced further for rejection and the algorithm is improved further by tightening the sigmoid functions' decision boundaries with Gaussian fitting. One possible drawback of this approach would be the lack of compact abating property of the sigmoids which may cause the problem of unbounded open space risk when they are activated by an infinitely distant input from all of the training data.\par
Since then, another research regarding document classification was reported by Shu et al.  \cite{shu2018unseen}. This classifier is the combining of a joint open classification with a sub-model. Seen and unseen classes are distinguished and rejected respectively by an open classification model. The sub-model finds the relation between two given samples to be identified if they belong to the same or different classes. Additionally, the number of invisible classes of the rejected samples can be obtained by considering this sub-model as a distance function for clustering.

\par
Another methodology for OSR is using a weightless neural network, denominated WiSARD \cite{aleksander2009brief}.
Compared to various classifiers, WiSARD does not rely on prior knowledge regarding data distribution, which is usually unavailable in OSR tasks. The proposed model assigns fitness scores to each class and evaluates how well a given observation matches the previously stored knowledge. This classifier applies such a fitting level for rejection according to the similarity rating and proximity between corresponding features. Computing score thresholds, this paper \cite{cardoso2015bounded} developed a rejection-capable WiSARD to identify whether observations pertaining to the class with the highest score or the best score is below the defined threshold, and then it is considered as an outlier. Following that, after proposing some exploratory results, a fully developed methodology is detailed in \cite{cardoso2017weightless}. This paper investigates how to adapt the WiSARD classifier for OSR by carrying out detailed distance-like calculations and defining the rejection thresholds at the training.
\par
Until recently, almost all existing deep open-set techniques included standard neural networks which are trained in a closed set environment and different activations which are analyzed to infer unknowns. However, relying on discriminative features of known classes in such systems causes specialization of learned representations to known classes and is not useful to represent unknowns. In contrast, some approaches enhance the learned representation to keep useful information to jointly perform known classification and unknown detection. Classification-Reconstruction learning for Open-Set Recognition (CROSR) is the novel framework proposed by Yoshihashi et al.  \cite{yoshihashi2019classification} most recently. This is the first neural network architecture which involved hierarchical reconstruction blocks and trained networks for joint classification and reconstruction of input samples. The proposed system consists of a closed-set classifier which exploits learned prediction $y$ for known class classification, and an unknown detector which uses a reconstructive latent representation $z$ together with $y$ for unknown detection, where $y$ and $z$ are provided by training a deep net. In this technique, utilizing reconstruction of input samples from low-dimensional latent representations \cite{hinton2006reducing}, allows unknown detectors to exploit a wider pool of features that may not be discriminative for known classes. This study which considers deep representation learning is similar to \cite{zhang2016sparse} in terms of sharing the idea of reconstruction-based representation learning; however, \cite{zhang2016sparse} uses a single layer linear representation.

\par
Recently, Oza et al. \cite{oza2019deep} combined a shared feature extractor that provides a latent space representation of an input image, along with a decoder and a classifier. While the decoder and classifier both take the latent representation as the input, the output of the decoder is the reconstructed image and that of classifier is the label of image. After training all networks and accomplishing both classification and reconstruction tasks, the reconstruction error tail from the known classes is modeled utilizing EVT to enhance the performance. Reconstruction errors from the decoder network are utilized to reject samples from the unknown classes. 
Another work \cite{oza2019c2ae} proposed an algorithm using class conditional auto-encoders. In this method, the training procedure is divided into two parts to improve the learning of open-set identification scores. The first part, closed-set classification, is learned by an encoder using the traditional classification loss and the closed-set training setting, while a decoder reconstructs conditioned on class identity to train an open-set identification model and accomplish the second part of the training. Furthermore, EVT is used to model reconstruction errors and obtain the operating threshold. 
\par
In a recent paper \cite{lian2019known}, the known and unknown classes are first distinguished based on entropy measurement and training the model on a modified cross entropy loss by dedicating a low and high cross-entropy for known and unknown classes, respectively. Then it uses the weighted square difference loss to assign unlabeled target samples to known classes based on the likeliness. Another work \cite{farfan2017towards} used CNN to extract effective features, along with a rejection approach depending on the uncertainty metric Breaking Ties \cite{luo2005active} to build a recognition method. During the rejection scenario, for a given test sample, the class confidence scores are computed. After that, the difference between the first and second best scores are used as an indicator to recognize an unknown sample. If this value goes over a pre-determined threshold, the observation is recognized as a known sample.

\section{Applications of Open Set Recognition}
In this section, we will make a general review of some practical applications of OSR. Emerging real-world recognition systems require OSR to recognize unknown inputs and learn them when needed. There is a multitude of real-world application domains where OSR can play a role, such as cyber-physical systems, intrusion recognition, face identification, video tracking and surveillance, image and text classification, spam filtering, forensics linguistics, movie genre classification, and document tagging \cite{battaglino2016open, krstulovic2018audio, scherreik2016multi, roos2017probabilistic, perera2017extreme, neira2018data, costa2014open, rocha2016authorship, navarro2018connecting, silver2013lifelong, schraml2017feasibility, poitevin2017challenges, bapst2017open, lampert2013attribute, chao2016empirical, xian2018zero, xian2017zero}. OSR is a challenging task in a large number of safety environments where even a small fraction of errors on unknowns could place human lives at risk, such as a self-driving car defect or robotic surgical assistants with flaws in perception and execution \cite{zamora2016novel, sunderhauf2016place, ramanagopal2018failing}. Moreover, real-world robots can expand their knowledge if they will be able to detect unknown objects, discover the need to learn about them and learn them continuously.
\par
An automatic face recognition system that is usually encountered with unknown individuals \cite{chiachia2014learning, pinto2015using, gunther2017unconstrained, bao2018towards, juefei2016multi, moeini2017open, gunther2017toward, jain2011handbook} is another domain to be deployed in open-universe scenarios. There is a wide range of real applications of face recognition, for instance, reducing retail crime, controlling mobile phone access, helping police officers, identifying people on social media platforms, and so on. This system consists of a feature extractor and a match component to do the face recognition. After feeding a face image into the system and extracting the biometric information, a match component compares the extracted features with the stored gallery faces. Face matching includes two different tasks: face verification and face identification. The face verification problem is to compare a pair of face pictures to decide whether the two face images represent the same individual or not. In the face identification, the comparison is against a gallery which contains a set of face images to recognize the corresponding identity of a given face picture. Although the face identification problem finds the nearest identity to the querying face, it can be treated as an open-set problem, and thus we need to decide whether the query subject is registered in the gallery or not. Initially, OSR is introduced by Li and Wechsler. \cite{li2005open} for face recognition task. There are a few works \cite{best2014unconstrained, stallkamp2007video, ekenel2009open, liao2014benchmark,sun2015face} which are mainly based on incorporating an operating threshold on similarity scores to address this problem. Other research \cite{wen2016discriminative, ortiz2014face, becker2013evaluating} points to the problem of face recognition with real-world databases in social media posts to determine and associate the most probable identity for the query face sample automatically.
\par
Another domain where OSR can be a solution is malware classification for cyber-security. Malware, shorthand for malicious software, meets the harmful intent of cyberattackers which is designed to pose severe and evolving security threats to individuals, government organizations, and private institutions. In the Internet age, with a higher frequency of communications among computer applications and their respective refinements, the number of new malware samples has explosively increased. Hence, it is required to keep up with the sophistication of newly received attacks and develop intelligent methods for effective and efficient malware detection. This domain is faced with the challenge of incomplete knowledge of the training data because of emerging novel types of malware. The ever-changing nature of malware, as the intruders are continuously altering network attacks to bypass the existing detection solutions, calls for the development of autonomous countermeasures and the recognition of novel malware classes \cite{henrydoss2017incremental, cruz2017open, hassen2018learning}. Rudd et al. surveyed many existing intrusion detection algorithms and proposed an open-world mathematical framework to extend and obviate the closed world assumption behind them \cite{rudd2016survey}. This flawed assumption impedes mappings between a machine learning solution and realistic malware recognition problems in which knowing all types of possible attacks cannot be known a priori. 
\par
Activity recognition has practical applications to facilitate human-vehicle communication and the transition to the level of driving automated systems. This task has the potential to recognize driver distraction for safety and improve dynamic driving adaptation like turning on the light if the person is reading a book or adjusting the seat while drinking coffee. However, it is difficult to apply computer vision models inside the vehicle cabin because of the dynamic nature of the surrounding environment. We cannot capture all possible driver behaviors in the training data, then the model, developed for closed set recognition, will be quickly exposed to uncertain situations and put the driver in disturbing and potentially dangerous situations. In \cite{roitberg2020open}, the task of open set driver activity recognition is introduced to address this issue. 

\section{Conclusion}
OSR arises due to an increased demand for a good classification or detection system, and thus a survey on this topic is particularly important. This survey tries to provide a structured and comprehensive overview of contemporary research on OSR. To the best of our knowledge, this is the first attempt to provide such a structured review that has been carried out in the field of OSR. Moreover, we reviewed rejection-adapted classifiers and anomaly/outlier techniques as the related tasks to OSR and analyzed their relationships. By comparing existing OSR techniques under two broad categories and discovering their limitations, we hope that this paper facilitates the promising subsequent research and a better understanding of this topic.\par
Based on the literature review, some related topics for future research can be discussed as follows.
As mentioned previously, the classification setting is usually a classic closed-set problem. In this setting, we face the risk of open space and misclassification of an unknown sample falling into over-occupied space divided for the known classes. An appropriate understanding of the nature and the underlying structure of the data can help us to arrange the known classes in a more compact form and limit the open space. The clustering technique is to create meaningful groups of the given samples based on the similarity that can improve the exploration of the data information and the generalization ability of classification learning. Thus, designing a learning framework that combines clustering and classification tasks can overcome the problem of over-occupied space. However, all current existing simultaneous learning clustering and classification algorithms are designed for closed-set problems. Therefore, designing such a framework under open-set assumptions could be a promising direction.
\par
In OSR, we do not know what all the classes are, so modeling unknown classes is not possible. One way to enhance the learning ability and robustness of the classifiers is by adopting the idea of adversarial learning. This novel technology is to generate examples that are close to the training data set but different from any training category. The fake data can be considered as unknown samples and used for the regularisation of the classifiers. The key factors that are worth further exploring are: how to generate valid examples, how to omit artificial selection by automatic construction of synthesis samples in the training process, what level of similarity should be designed between the distribution of synthesized examples and known samples. Although less similarity results in easier discrimination between known and synthesized samples, the performance of the classifier for unknown classes might be also lower. On the other hand, a high level of resemblance leads to achieving better discriminability during the training of the classifier. 
\par
Most of the OSR methods use threshold-based strategies in which the threshold is selected using the knowledge of the known classes. Thus, having no prior knowledge about unknown classes and defining a constant global threshold that no longer changes during the testing time, lead to OSR risk. That is, it misclassifies the unknown samples when they fall into the specified space for the known classes. Therefore, there is room to effectively determine a more robust way to select the threshold. For example, modifying the threshold based on the information of unknown classes received at the test time can improve the robustness of OSR techniques. Another investigation would be the robust selection of the tail's size while applying EVT to model the data distribution’s tail. This strategy can be helpful at the critical problems where outliers in known classes and unknown samples appear simultaneously in the testing phase. 
\par
Moreover, adversarial images are not necessarily isolated regions in the input space. They tend to stay adversarial across classifiers; consequently, they appear near a training sample and occupy contiguous areas in the input space. This problem is different from standard open space risk and presents a more difficult challenge towards existing OSR techniques, as they have not considered adversarial images in their experiments. It is, therefore, of great significance to design a framework for OSR that also takes the detection of adversarial images into account that can bring numerous benefits for the robustness and security of deep neural networks.

\bibliographystyle{unsrt}

\bibliography{cas-refs}

\end{document}